\documentclass[10pt,twocolumn,letterpaper]{article}

\usepackage{cvpr}
\usepackage{times}
\usepackage{epsfig}
\usepackage{graphicx}
\usepackage{amsmath}
\usepackage{amssymb}
\usepackage{subfigure}
\usepackage{multirow}
\usepackage[bottom]{footmisc}

\usepackage[pagebackref=true,breaklinks=true,letterpaper=true,colorlinks,bookmarks=false]{hyperref}

 \cvprfinalcopy 

\ifcvprfinal\fi
\begin{document}

\title{VoxelNet: End-to-End Learning for Point Cloud Based 3D Object Detection}

\author{Yin Zhou\\
Apple Inc\\
{\tt\small yzhou3@apple.com}
\and
Oncel Tuzel\\
Apple Inc\\
{\tt\small otuzel@apple.com}
}
\maketitle

\begin{abstract}
Accurate detection of objects in 3D point clouds is a central problem in many applications, such as autonomous navigation, housekeeping robots, and augmented/virtual reality. To interface a highly sparse LiDAR point cloud with a region proposal network (RPN), most existing efforts have focused on hand-crafted feature representations, for example, a bird's eye view projection. In this work, we remove the need of manual feature engineering for 3D point clouds and propose VoxelNet, a generic 3D detection network that unifies feature extraction and bounding box prediction into a single stage, end-to-end trainable deep network.
Specifically, VoxelNet divides a point cloud into equally spaced 3D voxels and transforms a group of points within each voxel into a unified feature representation through the newly introduced voxel feature encoding (VFE) layer. In this way, the point cloud is encoded as a descriptive volumetric representation, which is then connected to a RPN to generate detections. Experiments on the KITTI car detection benchmark show that VoxelNet outperforms  the state-of-the-art LiDAR based 3D detection methods by a large margin. Furthermore, our network learns an effective discriminative  representation of objects with various geometries, leading to encouraging results in 3D detection of pedestrians and cyclists, based on only LiDAR.

\end{abstract}

\section{Introduction}

\begin{figure}[!ht]
\centering
    \includegraphics[width=0.9\linewidth]{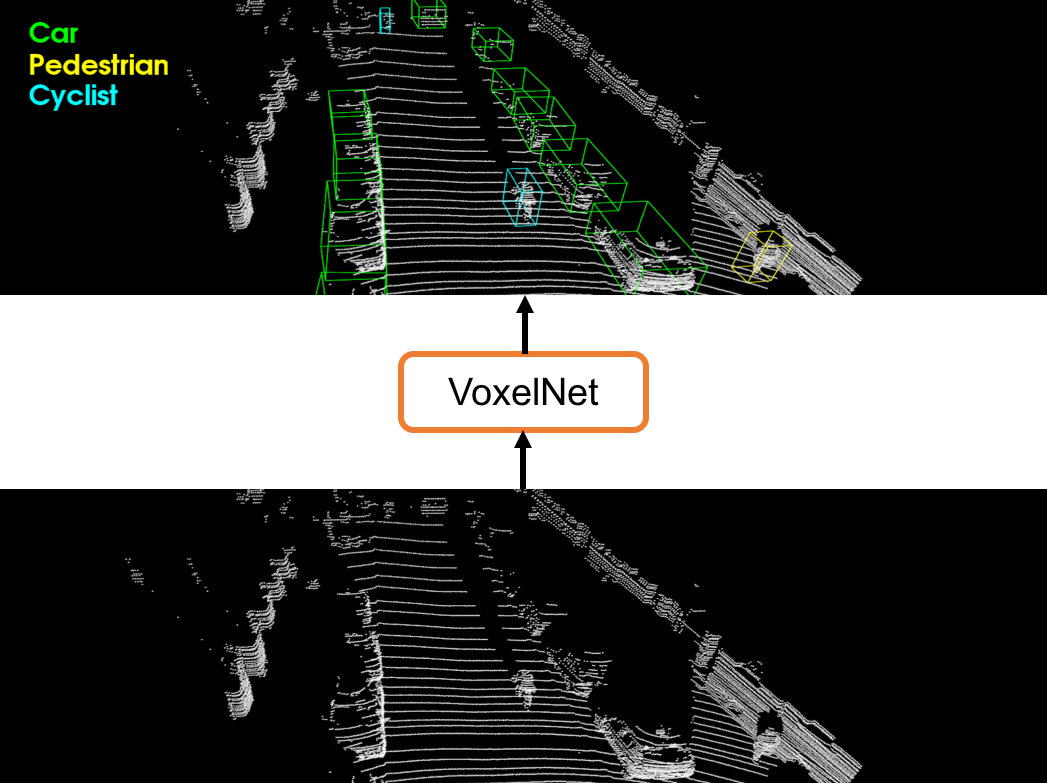}
\caption{VoxelNet directly operates on the raw point cloud (no need for feature engineering)  and produces the 3D detection results using a single end-to-end trainable network. }
\label{fig:1st_figure}
\end{figure}

\begin{figure*}[!th]
\centering
    \includegraphics[width=0.80\linewidth]{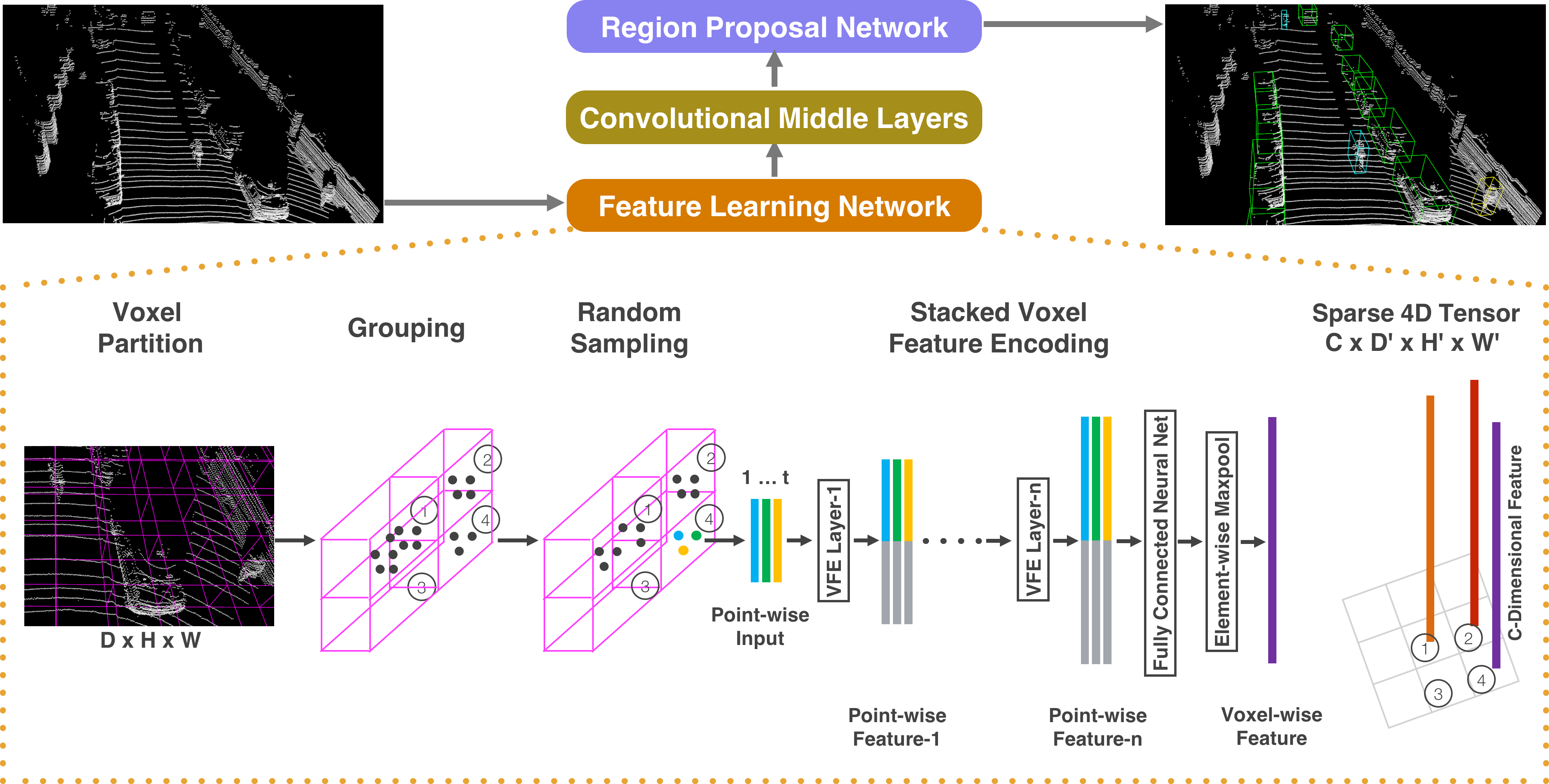}
    \vspace{-0.1cm}
\caption{VoxelNet architecture. The feature learning network takes a raw point cloud as input, partitions the space into voxels, and transforms points within each voxel to a vector representation characterizing the shape information. The space is represented as a sparse 4D tensor. The convolutional middle layers processes the 4D tensor to aggregate spatial context. Finally, a RPN generates the 3D detection.}
\label{fig:VoxelNet_flowchart}
\end{figure*}

Point cloud based 3D object detection is an important component of a variety of real-world applications, such as autonomous navigation~\cite{REF:Geiger2012CVPR,REF:Drone}, housekeeping robots~\cite{REF:Housekeeping}, and augmented/virtual reality~\cite{REF:AR}. Compared to image-based detection, LiDAR provides reliable depth information that can be used to accurately localize objects and characterize their shapes~\cite{REF:3DFCN, REF:cvpr17chen}. However, unlike images,  LiDAR point clouds are sparse and have highly variable point density, due to factors such as non-uniform sampling of the 3D space, effective range of the sensors, occlusion, and the relative pose. To handle these challenges, many approaches manually crafted feature representations for point clouds that are tuned for 3D object detection. Several methods project point clouds into a perspective view and apply image-based feature extraction techniques~\cite{REF:FusionDPM-IROS14, REF:MV-RGBD-RF2015,REF:VeloFCN}. Other approaches rasterize point clouds into a 3D voxel grid and encode each voxel with hand-crafted features~\cite{REF:Wang-RSS-15,REF:Vote3Deep,REF:Song2014,REF:DeepSlidingShapes,REF:3DFCN,REF:cvpr17chen}. However, these manual design choices introduce an information bottleneck that prevents these approaches from effectively exploiting 3D shape information and the required invariances for the detection task.  A major breakthrough in recognition~\cite{REF:NIPS2012_4824} and detection~\cite{REF:girshick2014rich} tasks on images was due to moving from hand-crafted features to machine-learned features.


Recently, Qi \textit{et al.}\cite{REF:qi2017pointnet} proposed PointNet, an end-to-end deep neural network that learns point-wise features directly from point clouds. This approach demonstrated impressive results on 3D object recognition, 3D object part segmentation, and point-wise semantic segmentation tasks. In \cite{REF:qi2017pointnetplusplus}, an improved version of PointNet was introduced which enabled the network to learn local structures at different scales. To achieve satisfactory results, these two approaches trained feature transformer networks on all input points ($\sim$1k points). Since typical point clouds obtained using LiDARs contain $\sim$100k points, training the architectures as in~\cite{REF:qi2017pointnet,REF:qi2017pointnetplusplus} results in high computational and memory requirements.   Scaling up 3D feature learning networks to orders of magnitude more points and to 3D detection tasks are the main challenges that we address in this paper. 


Region proposal network (RPN)~\cite{REF:NIPS2015_5638} is a highly optimized algorithm for efficient object detection~\cite{REF:ResNet2016,REF:cvpr17chen,REF:YOLO9000_2017,REF:SSD_Liu2016}. However, this approach requires data to be dense and organized in a tensor structure (e.g. image, video) which is not the case for typical LiDAR point clouds. In this paper, we close the gap between point set feature learning and RPN for 3D detection task.

We present VoxelNet, a generic 3D detection framework that simultaneously learns a discriminative feature representation from point clouds and predicts accurate 3D bounding boxes, in an end-to-end fashion, as shown in Figure~\ref{fig:VoxelNet_flowchart}.
We design a novel voxel feature encoding (VFE) layer, which enables inter-point interaction within a voxel, by combining point-wise features with a locally aggregated feature. Stacking multiple VFE layers allows learning complex features for characterizing local 3D shape information. Specifically, VoxelNet divides the point cloud into equally spaced 3D voxels, encodes each voxel via stacked VFE layers, and then 3D convolution further aggregates local voxel features, transforming the point cloud into a high-dimensional volumetric representation. Finally, a RPN consumes the volumetric representation and yields the detection result. This efficient algorithm benefits both from the sparse point structure and efficient parallel processing on the voxel grid.


We evaluate VoxelNet on the bird's eye view detection and the full 3D detection tasks, provided by the KITTI benchmark~\cite{REF:Geiger2012CVPR}. Experimental results show that VoxelNet outperforms  the state-of-the-art LiDAR based 3D detection methods by a large margin. We also demonstrate that VoxelNet achieves highly encouraging results in detecting pedestrians and cyclists from LiDAR point cloud.

\subsection{Related Work}
\label{sec:RelatedWork}


Rapid development of 3D sensor technology has motivated researchers to develop efficient representations to detect and localize objects in point clouds. Some of the earlier methods for feature representation are \cite{REF:StructuralIndexing_Medioni92,REF:COSMOS_Jain97,REF:PointSignatures_Chua1997,REF:SpinImageHebert99,REF:Tuzel2014,REF:FPFH_Beetz2009,REF:RoboticAssembly_Oncel2012,REF:Mian2010,REF:Nishino2010,REF:MS_POSE_Blake2011,REF:bo_iros11}. These hand-crafted features yield satisfactory results when rich and detailed 3D shape information is available. However their inability to adapt to more complex shapes and scenes, and learn required invariances from data resulted in limited success for uncontrolled scenarios such as autonomous navigation.

Given that images provide detailed texture information, many algorithms infered the 3D bounding boxes from 2D images~\cite{REF:nips15chen,REF:cvpr16chen,REF:xiang_cvpr15,REF:Zia2013,REF:Zia2014,REF:SFM2015}. However, the accuracy of  image-based 3D detection approaches are bounded by the accuracy of the depth estimation. 

Several LIDAR based 3D object detection techniques utilize a voxel grid representation.  \cite{REF:Wang-RSS-15, REF:Vote3Deep} encode each nonempty voxel with 6 statistical quantities that are derived from all the points contained within the voxel. \cite{REF:Song2014} fuses multiple local statistics to represent each voxel. \cite{REF:DeepSlidingShapes} computes the truncated signed distance on the voxel grid. \cite{REF:3DFCN} uses binary encoding for the 3D voxel grid. \cite{REF:cvpr17chen} introduces a multi-view representation for a LiDAR point cloud by computing a multi-channel feature map in the bird's eye view and the cylindral coordinates in the frontal view. Several other studies project point clouds onto a perspective view and then use image-based feature encoding schemes~\cite{REF:FusionDPM-IROS14, REF:MV-RGBD-RF2015,REF:VeloFCN}. 

There are also several multi-modal fusion methods that combine images and LiDAR to improve detection accuracy~\cite{REF:Enzweiler2011,REF:Gonzalez2017,REF:cvpr17chen}. These methods provide improved performance compared to LiDAR-only 3D detection, particularly for small objects (pedestrians, cyclists) or when the objects are far, since cameras provide an order of magnitude more measurements than LiDAR. However the need for an additional camera that is time synchronized and calibrated with the LiDAR restricts their use and makes the solution more sensitive to sensor failure modes. In this work we focus on LiDAR-only detection.

\subsection{Contributions}

\begin{itemize}
	\item  We propose a novel end-to-end trainable deep architecture for point-cloud-based 3D detection, VoxelNet, that directly operates on sparse 3D points and avoids information bottlenecks introduced by manual feature engineering.
	\item We present an efficient method to implement VoxelNet which benefits both from the sparse point structure and efficient parallel processing on the voxel grid.
	\item We conduct experiments on KITTI benchmark and show that VoxelNet produces state-of-the-art results in LiDAR-based car, pedestrian, and cyclist detection benchmarks.
\end{itemize}



\section{VoxelNet}
\label{sec:VoxelNet}
In this section we explain the architecture of VoxelNet, the loss function used for training, and an efficient algorithm to implement the network.  

\subsection{VoxelNet Architecture} 
The proposed VoxelNet consists of three functional blocks: (1) Feature learning network, (2) Convolutional middle layers, and (3) Region proposal network~\cite{REF:NIPS2015_5638}, as illustrated in Figure~\ref{fig:VoxelNet_flowchart}. We provide a detailed introduction of VoxelNet in the following sections.

\subsubsection{Feature Learning Network}
\label{subsec:Feature_Learning}
\noindent{\bf Voxel Partition } Given a point cloud, we subdivide the 3D space into equally spaced voxels as shown in Figure~\ref{fig:VoxelNet_flowchart}. Suppose the point cloud encompasses 3D space with range $D$, $H$, $W$ along the Z, Y, X axes respectively. We define each voxel of size $v_D$, $v_H$, and $v_W$ accordingly. The resulting 3D voxel grid is of size $D' = D/v_D, H' = H/v_H, W' = W/v_W$. Here, for simplicity, we assume $D$, $H$, $W$ are a multiple of  $v_D$, $v_H$, $v_W$. 

\noindent{\bf Grouping } We group the points according to the voxel they reside in. Due to factors such as distance, occlusion, object's relative pose, and non-uniform sampling, the LiDAR point cloud is sparse and has highly variable point density throughout the space. Therefore, after grouping, a voxel will contain a variable number of points. An illustration is shown in Figure~\ref{fig:VoxelNet_flowchart}, where Voxel-1 has significantly more points than Voxel-2 and Voxel-4, while Voxel-3  contains no point. 

\noindent{\bf Random Sampling } Typically a high-definition LiDAR point cloud is composed of $\sim$100k points. Directly processing all the points not only imposes increased memory/efficiency burdens on the computing platform, but also highly variable point density throughout the space might bias the detection. To this end, we  randomly sample a fixed number, $T$, of points from those voxels containing more than $T$ points. This sampling strategy has two purposes, (1) computational savings (see Section~\ref{subsec:implementation} for details); and (2) decreases the imbalance of points between the voxels which reduces the sampling bias, and adds more variation to training.

\begin{figure}[!t]
\centering
    \includegraphics[width=0.8\linewidth]{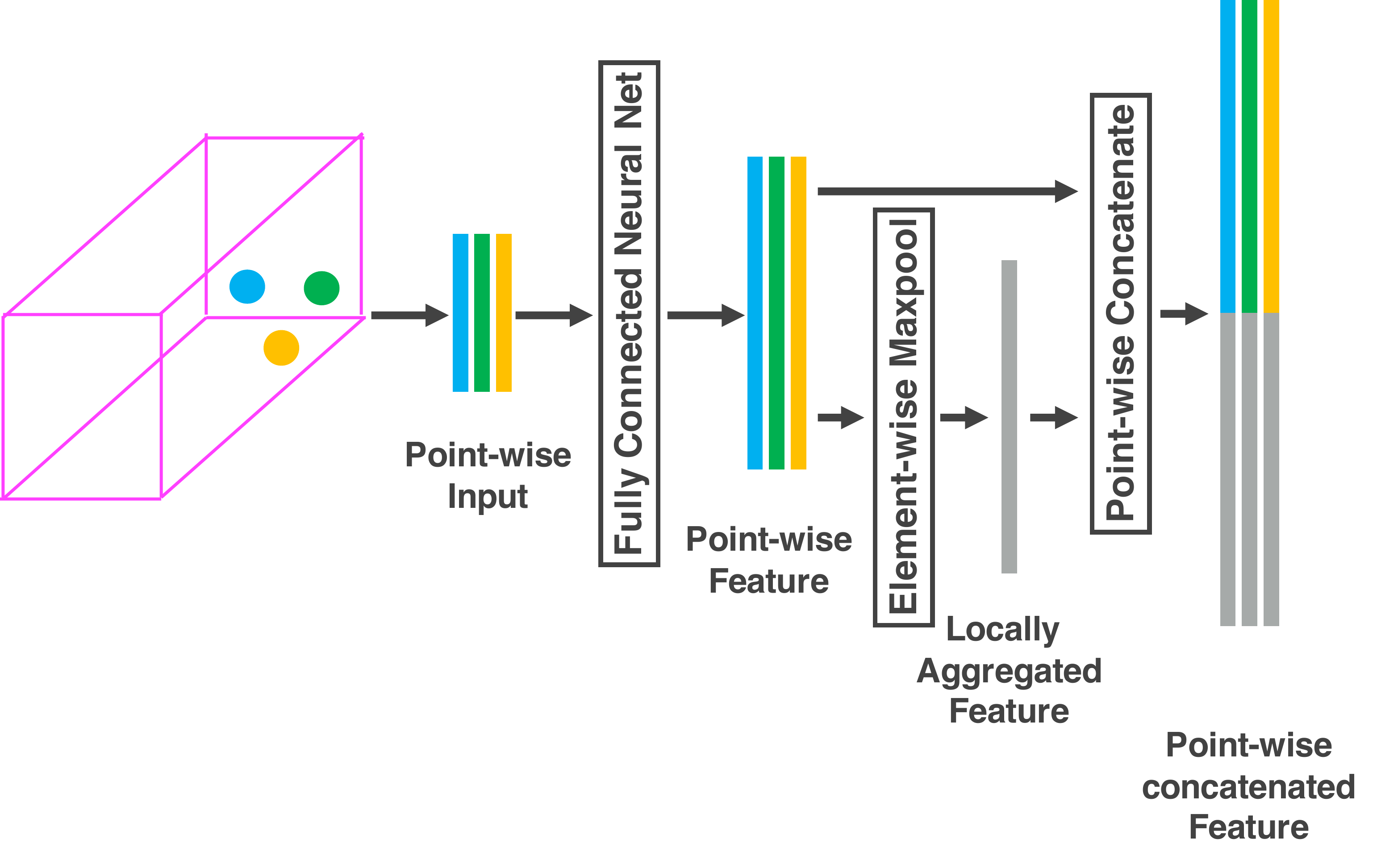}
      \vspace{-0.3cm}
\caption{Voxel feature encoding layer.}
\label{fig:VFE}
\end{figure}

\noindent{\bf Stacked Voxel Feature Encoding } The key innovation is the chain of VFE  layers. For simplicity, Figure~\ref{fig:VoxelNet_flowchart} illustrates the hierarchical feature encoding process for one voxel. Without loss of generality, we use VFE Layer-1 to describe the details in the following paragraph. Figure~\ref{fig:VFE} shows the architecture for VFE Layer-1. 

\begin{figure*}[!t]
\centering
    \includegraphics[width=0.8\linewidth]{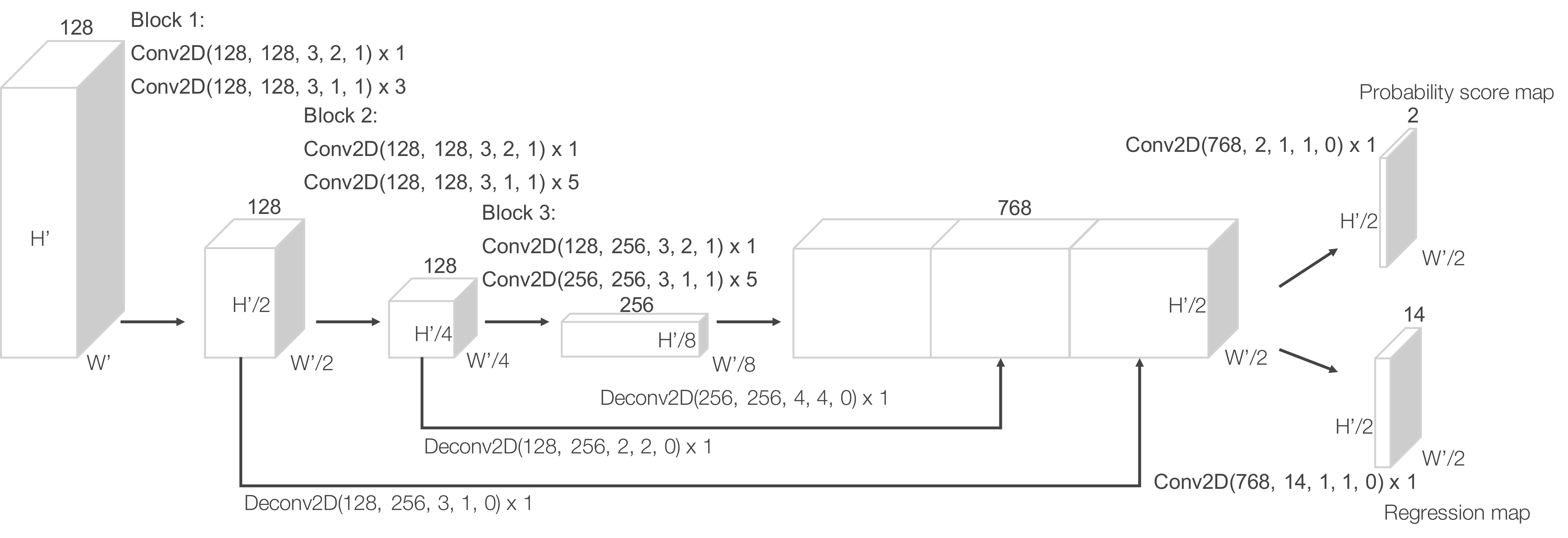}
    \vspace{-0.3cm}
\caption{Region proposal network architecture.}
\label{fig:RPN}
\end{figure*}

Denote $\mathbf{V} = \{\mathbf{p}_i=[x_i, y_i, z_i, r_i]^{T} \in \mathbb{R}^4\}_{i=1 \ldots t}$ as a non-empty voxel containing $t \leq T$ LiDAR points, where $\mathbf{p}_i$ contains XYZ coordinates for the $i$-th point and $r_i$ is the received reflectance. We first compute the local mean as the centroid of all the points in $\mathbf{V}$, denoted as $(v_x, v_y, v_z)$. Then we augment each point $\mathbf{p}_i$ with the relative offset w.r.t. the centroid  and obtain the input feature set $\mathbf{V}_{\textrm{in}} = \{ \hat{\mathbf{p}}_i=[x_i, y_i, z_i, r_i, x_i-v_x, y_i-v_y, z_i-v_z]^{T} \in \mathbb{R}^7 \}_{i=1 \ldots t}$. Next, each $\hat{\mathbf{p}}_i$ is transformed through the fully connected network (FCN) into a feature space, where we can aggregate information from the point features $\mathbf{f}_i \in \mathbb{R}^m$ to encode the shape of the surface contained within the voxel. The FCN is composed of a linear layer, a batch normalization (BN) layer, and a rectified linear unit (ReLU) layer. After obtaining point-wise feature representations, we use element-wise MaxPooling across all $\mathbf{f}_i$ associated to $\mathbf{V}$ to get the locally aggregated feature $\tilde{\mathbf{f}} \in \mathbb{R}^m$ for $\mathbf{V}$. Finally, we augment each $\mathbf{f}_i$ with $\tilde{\mathbf{f}}$ to form the point-wise concatenated feature as $\mathbf{f}_i^{out} = [\mathbf{f}_i^{T}, \tilde{\mathbf{f}}^{T}]^{T} \in \mathbb{R}^{2m}$. Thus we obtain the output feature set $\mathbf{V}_{\textrm{out}} = \{ \mathbf{f}_i^{out} \}_{i \ldots t}$. All non-empty voxels are encoded in the same way and they share the same set of parameters in FCN. 

We use $\textrm{VFE-i}(c_{in}, c_{out})$ to represent the $i$-th VFE layer that transforms input features of dimension $c_{in}$ into output features of dimension $c_{out}$. The linear layer learns a matrix of size $c_{in} \times (c_{out}/2)$, and the point-wise concatenation yields the output of dimension $c_{out}$.

Because the output feature combines both point-wise features and locally aggregated feature, stacking VFE layers encodes point interactions within a voxel and enables the final feature representation to learn descriptive shape information. The voxel-wise feature is obtained by transforming the output of VFE-$n$ into $\mathbb{R}^{C}$ via FCN and applying element-wise Maxpool where $C$ is the dimension of the voxel-wise feature, as shown in Figure~\ref{fig:VoxelNet_flowchart}.

\noindent{\bf Sparse Tensor Representation } By processing only the non-empty voxels, we obtain a list of voxel features, each uniquely associated to the spatial coordinates of a particular non-empty voxel. The obtained list of voxel-wise features can be represented as a sparse 4D tensor, of size $C \times D' \times H' \times W'$ as shown in Figure~\ref{fig:VoxelNet_flowchart}. Although the point cloud contains $\sim$100k points, more than $90\%$ of voxels typically are empty. Representing non-empty voxel features as a sparse tensor greatly reduces the memory usage and computation cost during backpropagation, and it is a critical step in our efficient implementation.



\subsubsection{Convolutional Middle Layers}
We use $\textrm{Conv}M\textrm{D}(c_{in}, c_{out}, \mathbf{k}, \mathbf{s}, \mathbf{p})$ to represent an $M$-dimensional convolution operator where $c_{in}$  and $c_{out}$ are the number of input and output channels, $\mathbf{k}$, $\mathbf{s}$, and $\mathbf{p}$ are the $M$-dimensional vectors corresponding to kernel size, stride size and padding size respectively. When the size across the $M$-dimensions are the same, we use a scalar to represent the size e.g. $k$ for $\mathbf{k} = (k,k,k)$. 

Each convolutional middle layer applies 3D convolution, BN layer, and ReLU layer sequentially. The convolutional middle layers aggregate voxel-wise features within a progressively expanding receptive field, adding more context to the shape description. The detailed sizes of the filters in the convolutional middle layers are explained in Section~\ref{sec:training_details}.


\subsubsection{Region Proposal Network}

 Recently, region proposal networks~\cite{REF:NIPS2015_5638} have become an important building block of top-performing object detection frameworks~\cite{REF:DeepSlidingShapes,REF:cvpr17chen,REF:focalloss}. In this work, we make several key modifications to the  RPN architecture proposed in~\cite{REF:NIPS2015_5638}, and combine it with the feature learning network and convolutional middle layers to form an end-to-end trainable pipeline. 

The input to our RPN is the feature map provided by the convolutional middle layers.  The architecture of this network is illustrated in Figure~\ref{fig:RPN}. The network has three blocks of fully convolutional layers. The first layer of each block downsamples the feature map by half via a convolution with a stride size of 2, followed by a sequence of convolutions of stride 1 ($\times q$ means $q$ applications of the filter). After each convolution layer, BN and ReLU operations are applied. We then upsample the output of every block to a fixed size and concatanate to construct the high resolution feature map. Finally, this feature map is mapped to the desired learning targets: (1) a probability score map and (2) a regression map.


\subsection{Loss Function}

Let $\{ a^{\text{\scriptsize pos}}_i \}_{i=1 \ldots N_{\text{\scriptsize pos}}}$ be the set of $N_{\text{\scriptsize pos}}$ positive anchors and $\{ a^{\text{\scriptsize neg}}_j \}_{j=1 \ldots N_{\text{\scriptsize neg}}}$ be the set of $N_{\text{\scriptsize neg}}$ negative anchors. We parameterize a 3D ground truth box as $(x_c^g, y_c^g, z_c^g, l^g, w^g, h^g, \theta^g)$, where $x_c^g, y_c^g, z_c^g$ represent the center location, $l^g, w^g, h^g$ are length, width, height of the box, and $\theta^g$ is the yaw rotation around Z-axis. To retrieve the ground truth box from a matching positive anchor parameterized as $(x_c^a, y_c^a, z_c^a, l^a, w^a, h^a, \theta^a)$, we define the residual vector $\mathbf{u}^* \in \mathbb{R}^7$  containing the 7 regression targets corresponding to center location $\Delta x, \Delta y, \Delta z$, three dimensions $\Delta l, \Delta w, \Delta h$, and the rotation $\Delta \theta$, which are computed as:
\begin{align}
\scriptsize
    &\Delta x = \frac{x_c^g - x_c^a}{d^a} , \Delta y = \frac{y_c^g - y_c^a}{d^a} , 
    \Delta z = \frac{z_c^g - z_c^a}{h^a} , \nonumber \\
    &\Delta l = \log(\frac{l^g}{l^a}) , \Delta w = \log(\frac{w^g}{w^a}) , \Delta h = \log(\frac{h^g}{h^a}) , \\
    &\Delta \theta = \theta^g - \theta^a \nonumber
\label{eqn:reg_targets}
\end{align}
where $d^a = \sqrt{(l^a)^2 + (w^a)^2}$ is the diagonal of the base of the anchor box. Here, we aim to directly estimate the oriented 3D box and normalize $\Delta x$ and $\Delta y$ homogeneously with the diagonal $d^a$, which is different from \cite{REF:NIPS2015_5638,REF:DeepSlidingShapes,REF:VeloFCN, REF:3DFCN,REF:nips15chen,REF:cvpr16chen,REF:cvpr17chen}. We define the loss function as follows:

\begin{eqnarray}
    L & = & \alpha\frac{1}{N_{\text{\scriptsize pos}}} \sum\limits_{i} L_{\text{\scriptsize cls}}(p^{\text{\scriptsize pos}}_i, 1) + 
            \beta\frac{1}{N_{\text{\scriptsize neg}}} \sum\limits_{j} L_{\text{\scriptsize cls}}(p^{\text{\scriptsize neg}}_j, 0) \nonumber \\
            & + & \frac{1}{N_{\text{\scriptsize pos}}} \sum\limits_{i} L_{\text{\scriptsize reg}}(\mathbf{u}_i, \mathbf{u}_i^*)
\label{eqn:loss_function}
\end{eqnarray}
where $p^{\text{\scriptsize pos}}_i$ and $p^{\text{\scriptsize neg}}_j$ represent the softmax output for positive anchor $a^{\text{\scriptsize pos}}_i$ and negative anchor $a^{\text{\scriptsize neg}}_j$ respectively, while $\mathbf{u}_i \in \mathbb{R}^7$ and $\mathbf{u}_i^* \in \mathbb{R}^7$ are the regression output and ground truth for positive anchor $a^{\text{\scriptsize pos}}_i$. The first two terms are the normalized classification loss for $\{ a^{\text{\scriptsize pos}}_i \}_{i=1 \ldots N_{\text{\scriptsize pos}}}$ and $\{ a^{\text{\scriptsize neg}}_j \}_{j=1 \ldots N_{\text{\scriptsize neg}}}$, where the $L_{\text{\scriptsize cls}}$ stands for binary cross entropy loss and $\alpha$, $\beta$ are postive constants balancing the relative importance. The last term $L_{reg}$ is the regression loss, where we use the SmoothL1 function~\cite{REF:Girshick:2015,REF:NIPS2015_5638}.


\begin{figure}[!t]
\centering
    \includegraphics[width=0.75\linewidth]{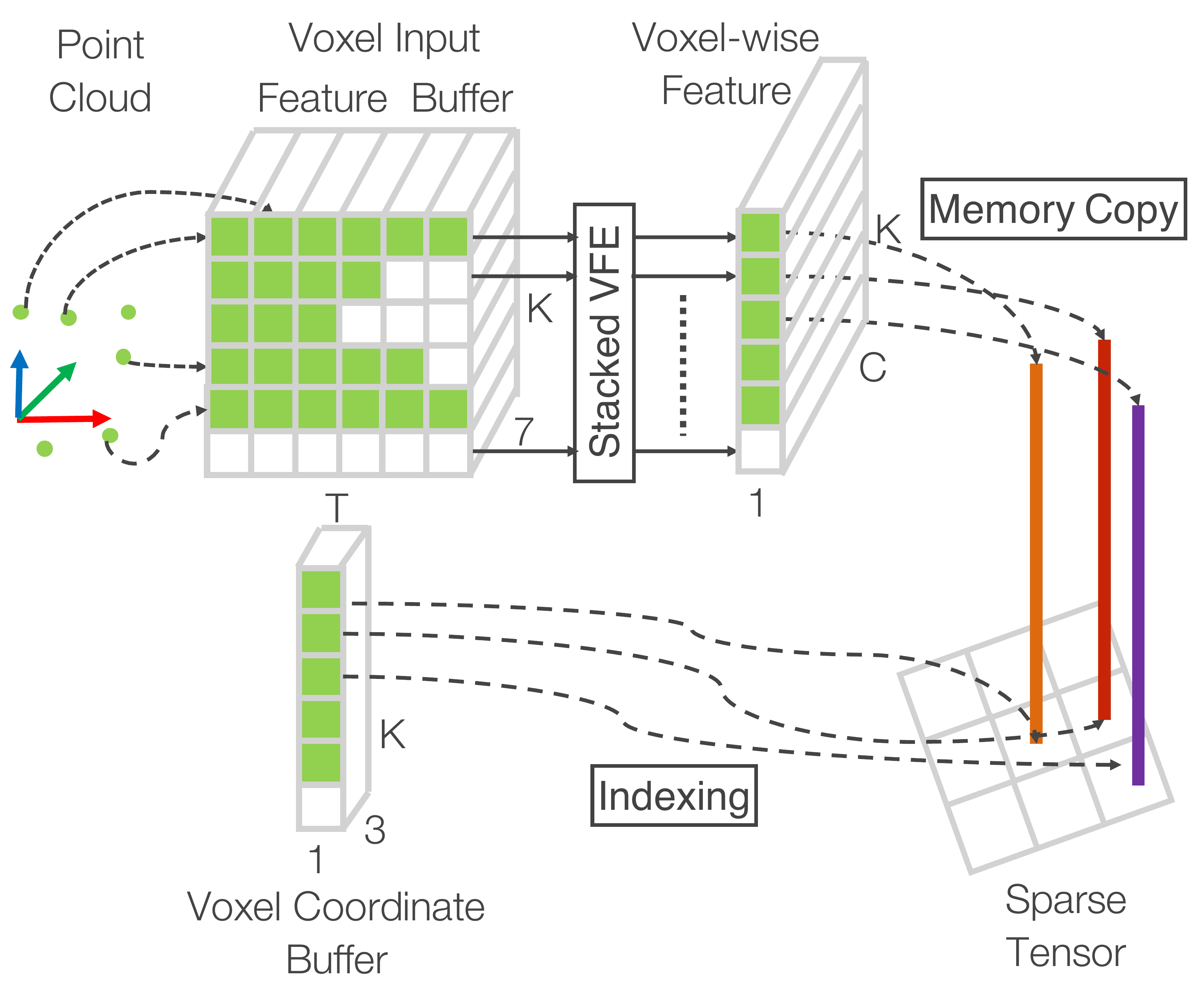}
\caption{Illustration of efficient implementation.}
\vspace{-0.3cm}
\label{fig:Implementation}
\end{figure}

\subsection{Efficient Implementation}
\label{subsec:implementation}

GPUs are optimized for processing dense tensor structures. The problem with working directly with the point cloud is that the points are sparsely distributed across space and each voxel has a variable number of points.  We devised a method that converts the point cloud into a dense tensor structure where stacked VFE operations can be processed in parallel  across points and voxels. 

The method is summarized in Figure~\ref{fig:Implementation}. We initialize a $K \times T \times 7$ dimensional tensor structure to store the voxel input feature buffer where $K$ is the maximum number of non-empty voxels, $T$ is the maximum number of points per voxel, and $7$ is the input encoding dimension for each point. The points are randomized before processing. For each point in the point cloud, we check if the corresponding voxel already exists. This lookup operation is done efficiently in $O(1)$ using a hash table where the voxel coordinate is used as the hash key. If the voxel is already initialized we insert the point to the voxel location if there are less than $T$ points, otherwise the point is ignored. If the voxel is not initialized, we initialize a new voxel, store its coordinate in the voxel coordinate buffer, and insert the point to this voxel location. The voxel input feature and coordinate buffers can be constructed via a single pass over the point list, therefore its complexity is $O(n)$.  To further improve the memory/compute efficiency it is possible to only store a limited number of voxels ($K$) and ignore points coming from voxels with few points. 

After the voxel input buffer is constructed, the stacked VFE only involves point level and voxel level dense operations which can be computed on a GPU in parallel. Note that,  after concatenation operations in VFE, we reset the features corresponding to empty points to zero such that they do not affect  the computed voxel features. Finally, using the stored coordinate buffer we reorganize the computed sparse voxel-wise structures to the dense voxel grid. The following  convolutional middle layers and RPN operations work on a dense voxel grid which can be efficiently implemented on a GPU.

\section{Training Details}
\label{sec:training_details}
In this section, we explain the implementation details of the VoxelNet and the training procedure. 
\subsection{Network Details} 
Our experimental setup is based on the LiDAR specifications of the KITTI dataset~\cite{REF:Geiger2012CVPR}.

\noindent{\bf Car Detection }  For this task, we consider point clouds within the range of $[-3, 1] \times [-40, 40] \times [0, 70.4]$ meters along Z, Y, X axis respectively. Points that are projected outside of image boundaries are removed~\cite{REF:cvpr17chen}. We choose a voxel size of $v_D=0.4, v_H=0.2, v_W=0.2$ meters, which leads to $D'=10$, $H'=400$, $W'=352$. We set $T=35$ as the maximum number of randomly sampled points in each non-empty voxel. We use two VFE layers $\textrm{VFE-1}(7, 32)$ and $\textrm{VFE-2}(32, 128)$. The final FCN maps VFE-2 output to $\mathbb{R}^{128}$. Thus our feature learning net generates a sparse tensor of shape $128 \times 10 \times 400 \times 352$. To aggregate voxel-wise features, we employ three convolution middle layers sequentially as Conv3D(128, 64, 3, (2,1,1), (1,1,1)), Conv3D(64, 64, 3, (1,1,1), (0,1,1)), and Conv3D(64, 64, 3, (2,1,1), (1,1,1)), which yields a 4D tensor of size $64 \times 2 \times 400 \times 352$. After reshaping, the input to RPN is a feature map of size $128 \times 400 \times 352$, where the dimensions correspond to channel, height, and width of the 3D tensor. Figure~\ref{fig:RPN} illustrates the detailed network architecture for this task. Unlike \cite{REF:cvpr17chen}, we use only one anchor size, $l^a=3.9, w^a=1.6, h^a=1.56$ meters, centered at $z_c^a = -1.0$ meters with two rotations, 0 and 90 degrees. 
Our anchor matching criteria is as follows: An anchor is considered as positive if it has the highest Intersection over Union (IoU) with a ground truth or its IoU with ground truth is above 0.6 (in bird's eye view). An anchor is considered as negative if the IoU between it and all ground truth boxes is less than 0.45. We treat anchors as don't care if they have $0.45 \leq \textrm{IoU} \leq 0.6$ with any ground truth. We set $\alpha = 1.5$ and $\beta = 1$ in Eqn.~\ref{eqn:loss_function}.

\noindent{\bf Pedestrian and Cyclist Detection } The input range\footnote{Our empirical observation suggests that beyond this range, LiDAR returns from pedestrians and cyclists become very sparse and therefore detection results will be unreliable.} is $[-3, 1] \times [-20, 20] \times [0, 48]$ meters along Z, Y, X axis respectively. We use the same voxel size as for car detection, which yields $D=10$, $H=200$, $W=240$. We set $T=45$ in order to obtain more LiDAR points for better capturing shape information. The feature learning network and convolutional middle layers are identical to the networks used in the car detection task. For the RPN, we make one modification to block 1 in Figure~\ref{fig:RPN} by changing the stride size in the first 2D convolution from 2 to 1. This allows finer resolution in anchor matching, which is necessary for detecting pedestrians and cyclists. We use anchor size $l^a=0.8, w^a=0.6, h^a=1.73$ meters  centered at $z_c^a = -0.6$ meters with 0 and 90 degrees rotation for pedestrian detection and use anchor size $l^a=1.76, w^a=0.6, h^a=1.73$ meters centered at $z_c^a = -0.6$ with 0 and 90 degrees rotation for cyclist detection.
The specific anchor matching criteria is as follows: We assign an anchor as postive if it has the highest IoU with a ground truth, or its IoU with ground truth is above 0.5. An anchor is considered as negative if its IoU with every ground truth is less than 0.35. For anchors having $0.35 \leq \textrm{IoU} \leq 0.5$ with any ground truth, we treat them as don't care. 

During training, we use stochastic gradient descent (SGD) with learning rate 0.01 for the first 150 epochs and decrease the learning rate to 0.001 for the last 10 epochs. We use a batchsize of 16 point clouds.

\subsection{Data Augmentation}
With less than 4000 training point clouds, training our network from scratch will inevitably suffer from overfitting. To reduce this issue, we introduce three different forms of data augmentation. The augmented training data are generated on-the-fly without the need to be stored on disk~\cite{REF:NIPS2012_4824}. 

Define set $\mathbf{M} = \{ \mathbf{p}_i=[x_i, y_i, z_i, r_i]^T \in \mathbb{R}^4 \}_{i=1, \ldots, N}$ as the whole point cloud, consisting of $N$ points.
We parameterize a  3D bouding box $\mathbf{b}_i$ as $(x_c, y_c, z_c, l, w, h, \theta)$, where $x_c, y_c, z_c$ are center locations, $l, w, h$ are length, width, height, and $\theta$ is the yaw rotation around Z-axis. We define $\Omega_i=\{\mathbf{p} | x \in [x_c - l/2, x_c + l/2], y \in [y_c - w/2, y_c + w/2], z \in [z_c - h/2, z_c + h/2], \mathbf{p} \in \mathbf{M}\}$ as the set containing all LiDAR points within $\mathbf{b}_i$, where $\mathbf{p}=[x, y, z, r]$ denotes a particular LiDAR point in the whole set $\mathbf{M}$.

The first form of data augmentation applies perturbation independently to each ground truth 3D bounding box together with those LiDAR points within the box. Specifically, around Z-axis we rotate $\mathbf{b}_i$ and the associated $\Omega_i$ with respect to $(x_c, y_c, z_c)$ by a uniformally distributed random variable $\Delta\theta \in [-\pi/10, +\pi/10]$. Then we add a translation $(\Delta x, \Delta y, \Delta z)$ to the XYZ components of $\mathbf{b}_i$ and to each point in $\Omega_i$, where $\Delta x$, $\Delta y$, $\Delta z$ are drawn independently from a Gaussian distribution with mean zero and standard deviation 1.0. To avoid physically impossible outcomes, we perform a collision test between any two boxes after the perturbation and revert to the original if a collision is detected. Since the perturbation is applied to each ground truth box and the associated LiDAR points independently, the network is able to learn from substantially more variations than from the original training data.

Secondly, we apply global scaling to all ground truth boxes $\mathbf{b}_i$ and to the whole point cloud $\mathbf{M}$. Specifically, we multiply the XYZ coordinates and the three dimensions of each $\mathbf{b}_i$, and the XYZ coordinates of all points in $\mathbf{M}$ with a random variable drawn from uniform distribution $[0.95, 1.05]$. Introducing global scale augmentation improves robustness of the network for detecting objects with various sizes and distances as shown in image-based classification~\cite{REF:Simonyan14c, REF:journals/corr/Howard13} and detection tasks~\cite{REF:Girshick:2015, REF:ResNet2016}.

Finally, we apply global rotation to all ground truth boxes $\mathbf{b}_i$ and to the whole point cloud $\mathbf{M}$. The rotation is applied along Z-axis and around $(0, 0, 0)$. The global rotation offset is determined by sampling from uniform distribution $[-\pi/4, +\pi/4]$. By rotating the entire point cloud, we simulate the vehicle making a turn.


\begin{table*}[!t]
\centering
\scalebox{0.9}{
\begin{tabular}{|c|c||c|c|c||c|c|c||c|c|c|}
\hline
\multirow{2}{*}{Method} & \multirow{2}{*}{Modality} & \multicolumn{3}{c||}{Car} & \multicolumn{3}{c||}{Pedestrian} & \multicolumn{3}{c|}{Cyclist} \\\cline{3-11}
     & & Easy & Moderate & Hard & Easy & Moderate & Hard & Easy & Moderate & Hard  \\
\hline
Mono3D~\cite{REF:cvpr16chen} & Mono & 5.22 & 5.19 & 4.13 & N/A & N/A & N/A & N/A & N/A & N/A \\ 

3DOP~\cite{REF:nips15chen} & Stereo & 12.63 & 9.49 & 7.59 & N/A & N/A & N/A & N/A & N/A & N/A \\

VeloFCN~\cite{REF:VeloFCN} & LiDAR & 40.14 & 32.08 & 30.47 & N/A & N/A & N/A & N/A & N/A & N/A \\

MV (BV+FV)~\cite{REF:cvpr17chen} & LiDAR & 86.18 & 77.32 & 76.33 & N/A & N/A & N/A & N/A & N/A & N/A \\

MV (BV+FV+RGB)~\cite{REF:cvpr17chen} & LiDAR+Mono & 86.55 & 78.10 & 76.67 & N/A & N/A & N/A & N/A & N/A & N/A \\
\hline
HC-baseline & LiDAR & 88.26 & 78.42 & 77.66  & 58.96 & 53.79 & 51.47 & 63.63 & 42.75 & 41.06 \\

VoxelNet & LiDAR & \bf{89.60} & \bf{84.81} & \bf{78.57} & \bf{65.95} & \bf{61.05} & \bf{56.98} & \bf{74.41} & \bf{52.18} & \bf{50.49} \\
\hline
\end{tabular}
}
\caption{Performance comparison in bird's eye view detection: average precision (in \%) on KITTI validation set.}
\vspace{-0.3cm}

\label{table:BEV_detection_comparison}
\end{table*}

\begin{table*}[!t]
\centering
\scalebox{0.9}{
\begin{tabular}{|c|c||c|c|c||c|c|c||c|c|c|}
\hline
\multirow{2}{*}{Method} & \multirow{2}{*}{Modality} & \multicolumn{3}{c||}{Car} & \multicolumn{3}{c||}{Pedestrian} & \multicolumn{3}{c|}{Cyclist} \\\cline{3-11}
     & & Easy & Moderate & Hard & Easy & Moderate & Hard & Easy & Moderate & Hard  \\
\hline
Mono3D~\cite{REF:cvpr16chen} & Mono & 2.53 & 2.31 & 2.31 & N/A & N/A & N/A & N/A & N/A & N/A \\ 

3DOP~\cite{REF:nips15chen} & Stereo & 6.55 & 5.07 & 4.10 & N/A & N/A & N/A & N/A & N/A & N/A \\

VeloFCN~\cite{REF:VeloFCN} & LiDAR & 15.20 & 13.66 & 15.98 & N/A & N/A & N/A & N/A & N/A & N/A \\

MV (BV+FV)~\cite{REF:cvpr17chen} & LiDAR & 71.19 & 56.60 & 55.30 & N/A & N/A & N/A & N/A & N/A & N/A \\

MV (BV+FV+RGB)~\cite{REF:cvpr17chen} & LiDAR+Mono & 71.29 & 62.68 & 56.56 & N/A & N/A & N/A & N/A & N/A & N/A \\
\hline
HC-baseline & LiDAR & 71.73 & 59.75 & 55.69 & 43.95 & 40.18 & 37.48 & 55.35 & 36.07 & 34.15 \\

VoxelNet & LiDAR & \bf{81.97} & \bf{65.46} & \bf{62.85} & \bf{57.86} & \bf{53.42} & \bf{48.87} & \bf{67.17} & \bf{47.65} & \bf{45.11} \\
\hline
\end{tabular}
}
\caption{Performance comparison in 3D detection: average precision (in \%) on KITTI validation set.}
\vspace{-0.1cm}
\label{table:3D_detection_comparison}
\end{table*}

\section{Experiments}
\label{sec:Experiment}
We evaluate VoxelNet on the KITTI 3D object detection benchmark~\cite{REF:Geiger2012CVPR} which contains 7,481 training images/point clouds and 7,518 test images/point clouds, covering three categories: \textit{Car}, \textit{Pedestrian}, and \textit{Cyclist}. For each class, detection outcomes are evaluated based on three difficulty levels: \textit{easy}, \textit{moderate}, and \textit{hard}, which are determined according to the object size, occlusion state, and truncation level. Since the ground truth for the test set is not available and the access to the test server is limited, we conduct comprehensive evaluation using the protocol described in~\cite{REF:nips15chen,REF:cvpr16chen,REF:cvpr17chen} and subdivide the training data into a training set and a validation set, which results in 3,712 data samples for training and 3,769 data samples for validation. The split avoids samples from the same sequence being included in both the training and the validation set~\cite{REF:cvpr16chen}. Finally we also present the test results using the KITTI server.

For the \textit{Car} category, we compare the proposed method with several top-performing algorithms, including image based approaches: Mono3D~\cite{REF:cvpr16chen} and 3DOP~\cite{REF:nips15chen}; LiDAR based approaches: VeloFCN~\cite{REF:VeloFCN} and 3D-FCN~\cite{REF:3DFCN}; and a multi-modal approach MV~\cite{REF:cvpr17chen}. Mono3D~\cite{REF:cvpr16chen}, 3DOP~\cite{REF:nips15chen} and MV~\cite{REF:cvpr17chen}  use a pre-trained model for initialization whereas we train VoxelNet from scratch using only the LiDAR data provided in KITTI.

To analyze the importance of end-to-end learning, we implement a strong baseline that is derived from the VoxelNet architecture but uses hand-crafted features instead of the proposed feature learning network. We call this model the hand-crafted baseline (HC-baseline). HC-baseline uses the bird's eye view features described in 
\cite{REF:cvpr17chen} which are computed at $0.1$m resolution. Different from \cite{REF:cvpr17chen}, we increase the number of height channels from 4 to 16 to capture more detailed shape information-- further increasing the number of height channels did not lead to performance improvement. We replace the convolutional middle layers of VoxelNet with similar size 2D convolutional layers, which are Conv2D(16, 32, 3, 1, 1), Conv2D(32, 64, 3, 2, 1), Conv2D(64, 128, 3, 1, 1). Finally RPN is identical in VoxelNet and HC-baseline. The total number of parameters in HC-baseline and VoxelNet are very similar. We train the HC-baseline using the same training procedure and data augmentation described in Section~\ref{sec:training_details}.

\begin{figure*}[!t]
\centering
    \begin{tabular}{ccc}
    \centering
   
\vspace{-0.30cm}
        \hspace{-0.75cm} \includegraphics[width=0.38\linewidth]{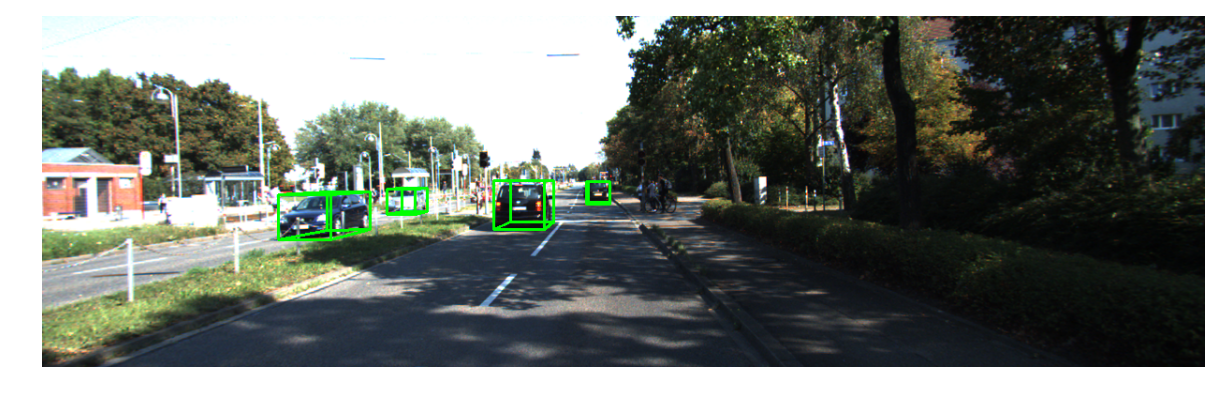} & \hspace{-1.25cm}

        \includegraphics[width=0.38\linewidth]{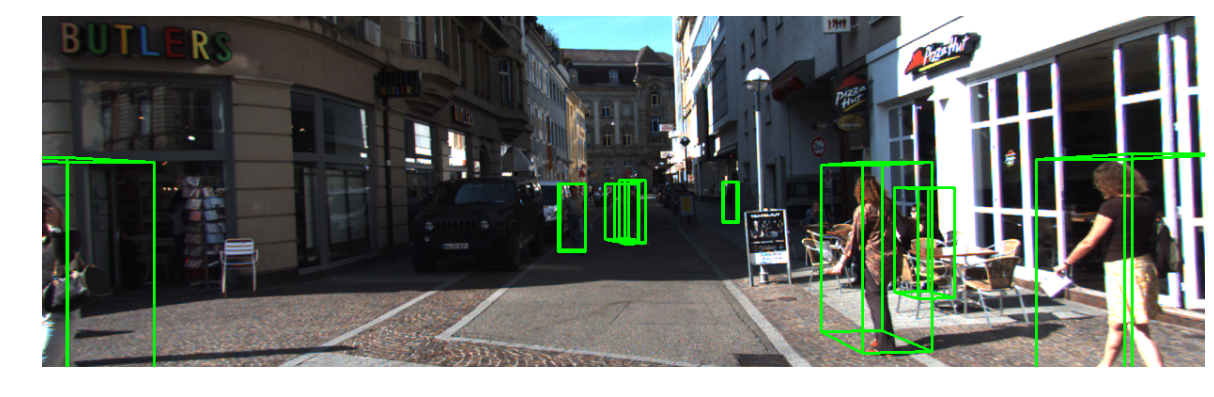} & \hspace{-1.25cm}

        \includegraphics[width=0.38\linewidth]{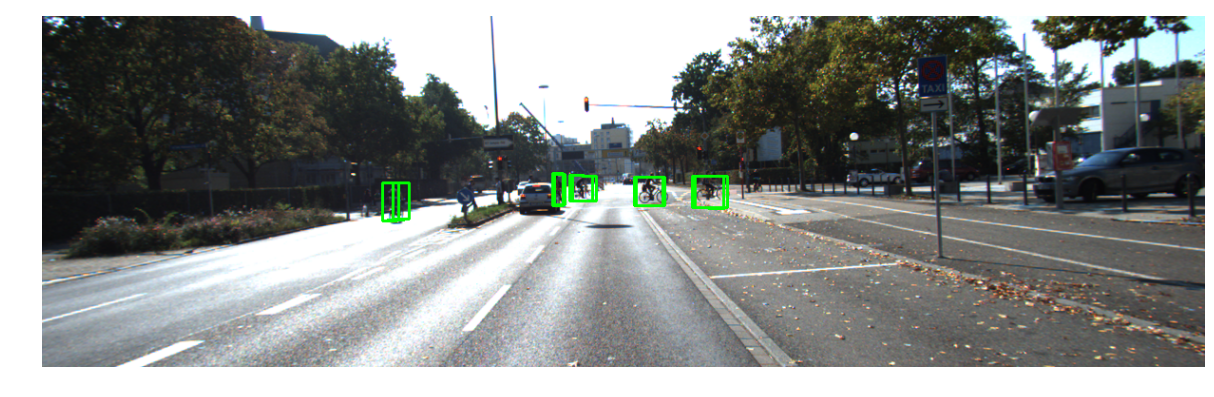} \\

\vspace{-0.30cm}

        \hspace{-0.75cm} \includegraphics[width=0.38\linewidth]{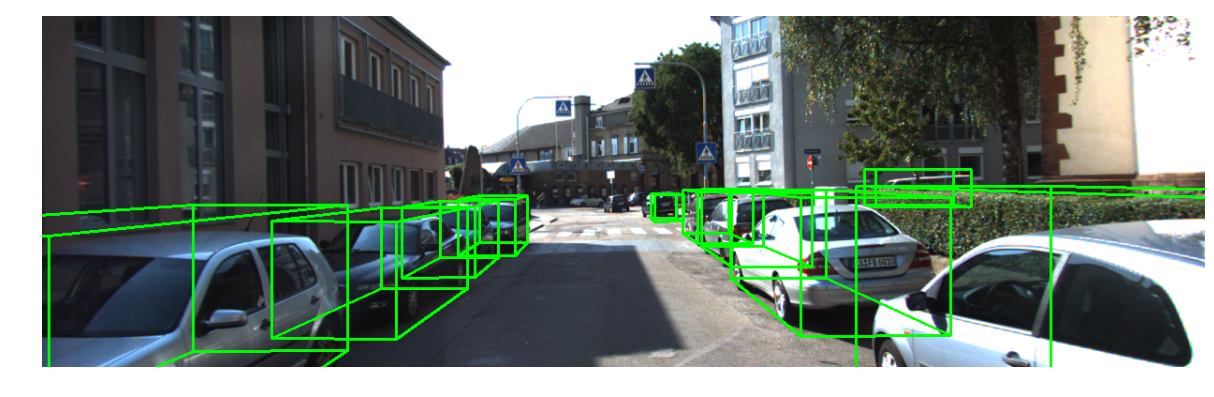} & \hspace{-1.25cm}
        \includegraphics[width=0.38\linewidth]{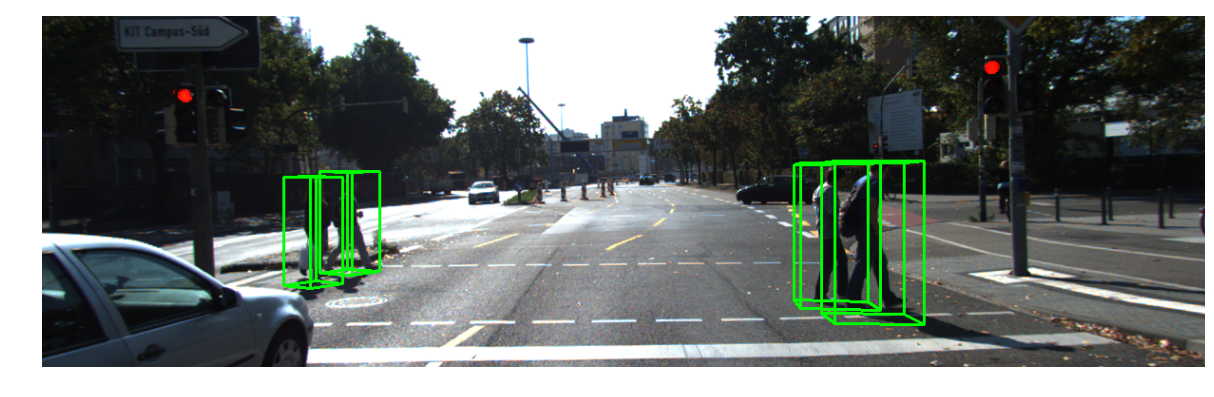} & \hspace{-1.25cm}
        \includegraphics[width=0.38\linewidth]{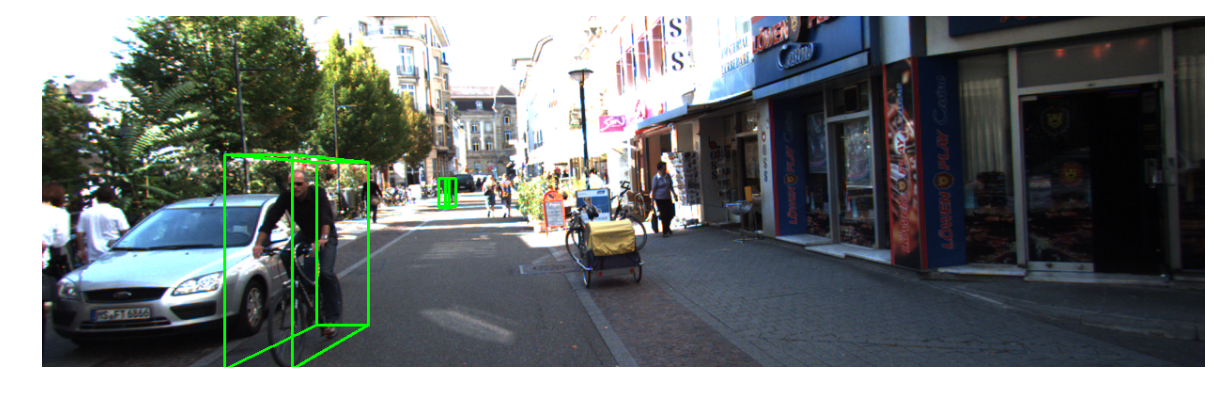} \\

\vspace{-0.30cm}

        \hspace{-0.75cm} \includegraphics[width=0.38\linewidth]{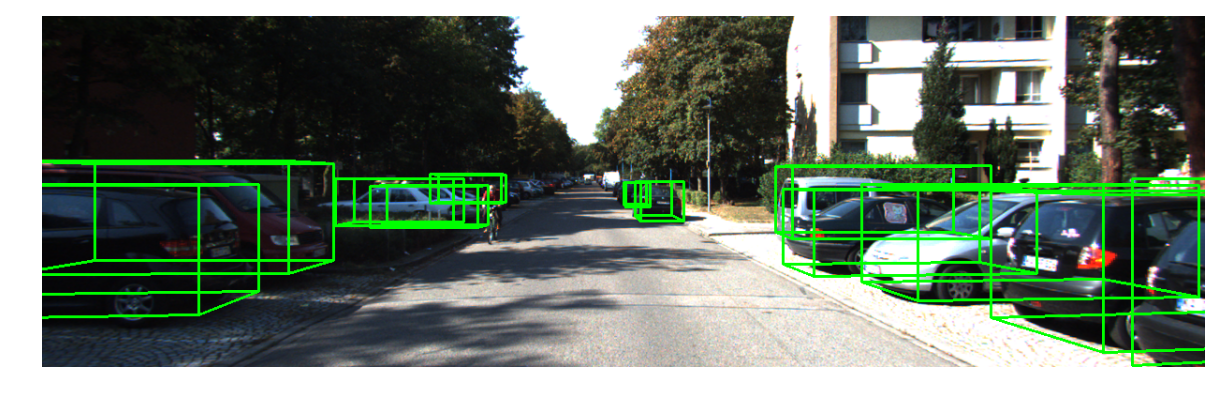} & \hspace{-1.25cm}
        \includegraphics[width=0.38\linewidth]{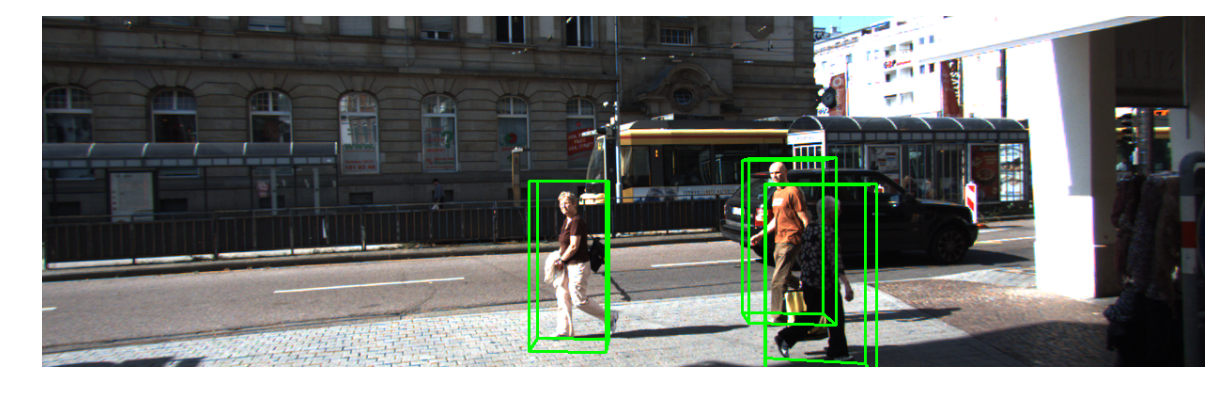} & \hspace{-1.25cm}
        \includegraphics[width=0.38\linewidth]{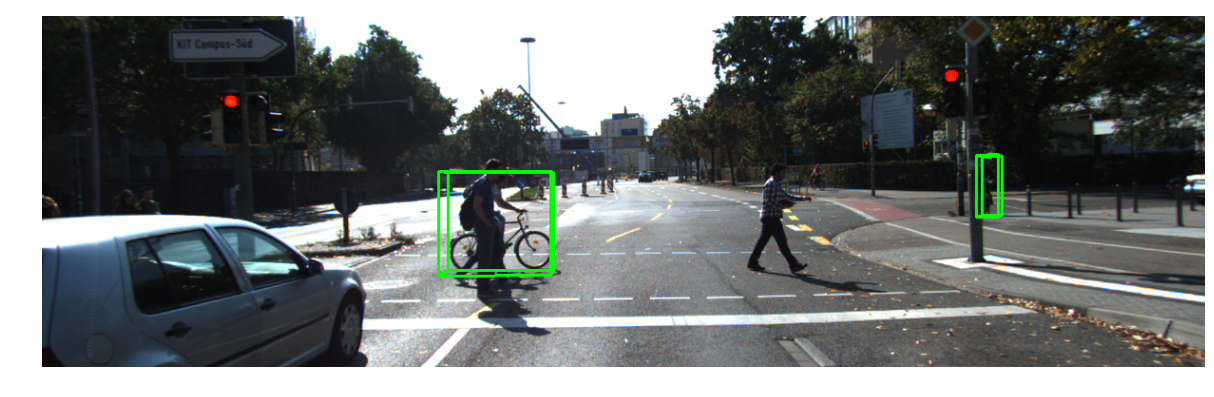} \\

\vspace{-0.30cm}

        \hspace{-0.75cm} \includegraphics[width=0.38\linewidth]{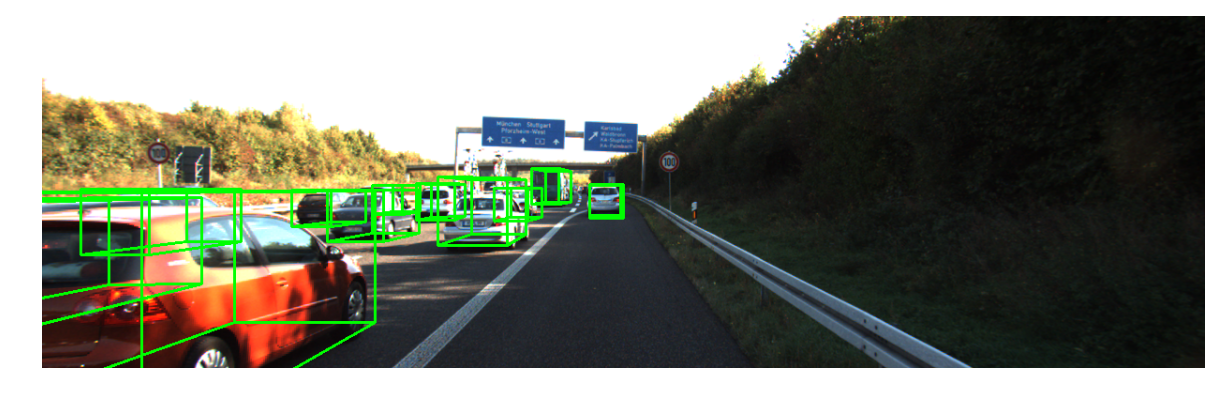} & \hspace{-1.25cm}
        \includegraphics[width=0.38\linewidth]{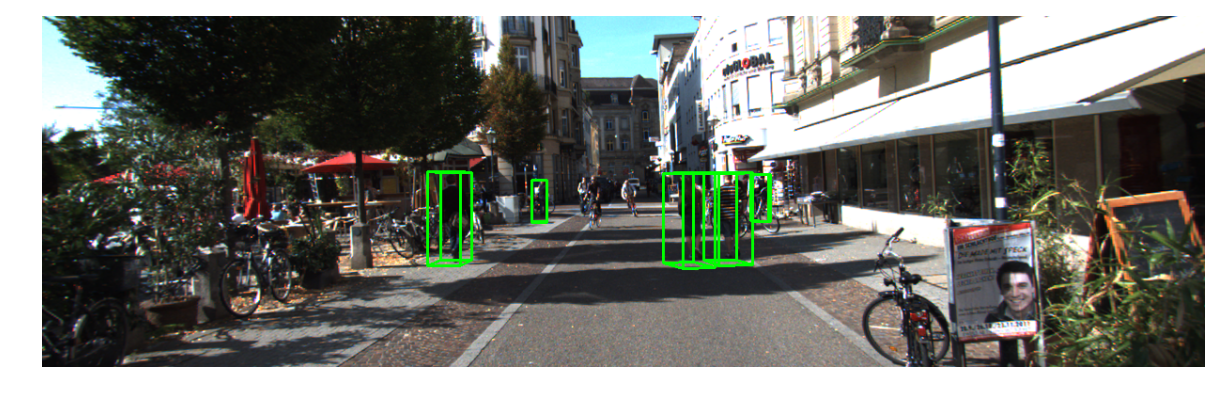} & \hspace{-1.25cm}
        \includegraphics[width=0.38\linewidth]{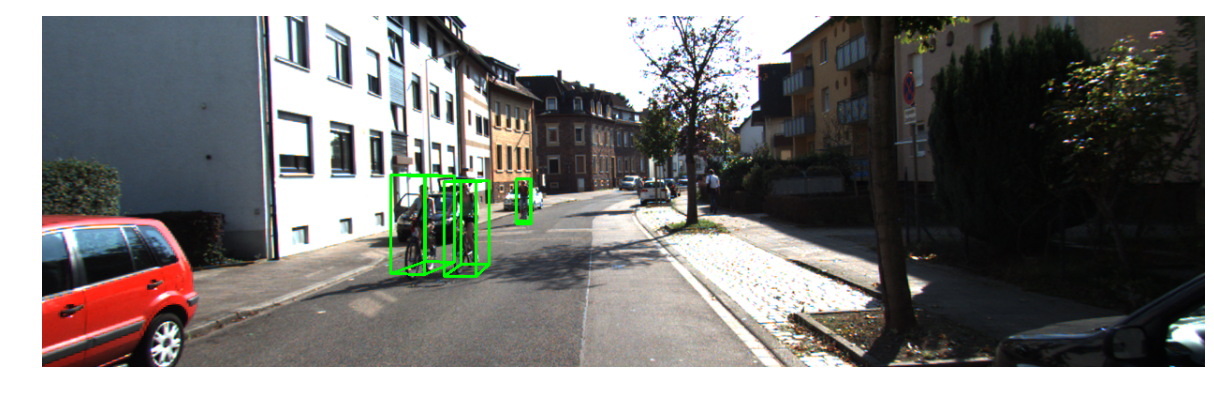} \\

        \hspace{-0.75cm} Car & \hspace{-1.25cm} Pedestrian & \hspace{-1.25cm} Cyclist \\
    \end{tabular}
    \caption{Qualitative results. For better visualization  3D boxes detected using LiDAR are projected on to the RGB images.}
    \label{fig:Qualitative_demo}
\end{figure*}

\subsection{Evaluation on KITTI Validation Set}
\paragraph{Metrics} 
We follow the official KITTI evaluation protocol, where the IoU threshold is 0.7 for class \textit{Car} and is 0.5 for class \textit{Pedestrian} and \textit{Cyclist}. The IoU threshold is the same for both bird's eye view and full 3D evaluation. We compare the methods using the average precision (AP) metric.

\paragraph{Evaluation in Bird's Eye View} 

The evaluation result is presented in Table~\ref{table:BEV_detection_comparison}. VoxelNet consistently outperforms all the competing approaches across all three difficulty levels. HC-baseline also achieves satisfactory performance compared to the state-of-the-art~\cite{REF:cvpr17chen}, which shows that our base region proposal network (RPN) is effective. For \textit{Pedestrian} and \textit{Cyclist} detection tasks in bird's eye view, we compare the proposed VoxelNet with HC-baseline. VoxelNet yields substantially higher AP than the HC-baseline for these more challenging categories, which shows that end-to-end learning is essential for point-cloud based detection.

We would like to note that \cite{REF:3DFCN} reported 88.9\%, 77.3\%, and 72.7\% for \textit{easy}, \textit{moderate}, and \textit{hard} levels respectively, but these results are obtained based on a different split of 6,000 training frames and $\sim$1,500 validation frames, and they are not directly comparable with algorithms in Table~\ref{table:BEV_detection_comparison}. Therefore, we do not include these results in the table.

\paragraph{Evaluation in 3D}
Compared to the bird's eye view detection, which requires only accurate localization of objects in the 2D plane, 3D detection is a more challenging task as it requires finer localization of shapes in 3D space. Table~\ref{table:3D_detection_comparison} summarizes the  comparison. For the class \textit{Car}, VoxelNet significantly outperforms all other approaches in AP across all difficulty levels. Specifically, using only LiDAR, VoxelNet significantly outperforms the state-of-the-art method MV (BV+FV+RGB)~\cite{REF:cvpr17chen} based on LiDAR+RGB, by 10.68\%, 2.78\% and 6.29\% in \textit{easy}, \textit{moderate}, and \textit{hard} levels respectively. HC-baseline achieves similar accuracy to the MV~\cite{REF:cvpr17chen} method.

As in the bird's eye view evaluation, we also compare VoxelNet with HC-baseline on 3D \textit{Pedestrian} and \textit{Cyclist} detection. Due to the high variation in 3D poses and shapes, successful detection of these two categories requires better 3D shape representation. As shown in Table~\ref{table:3D_detection_comparison} the improved performance of VoxelNet is emphasized for more challenging 3D detection tasks (from $\sim$8\% improvement in bird's eye view to $\sim$12\% improvement on 3D detection) which suggests that VoxelNet is more effective in capturing 3D shape information than hand-crafted features.

\subsection{Evaluation on KITTI Test Set}

We evaluated VoxelNet on the KITTI test set by submitting detection results to the official server. The results are summarized in Table~\ref{table:test_set_result}. VoxelNet,  significantly outperforms the previously published state-of-the-art~\cite{REF:cvpr17chen}  in all the tasks (bird's eye view and 3D detection) and all difficulties. We would like to note that many of the other leading methods listed in KITTI benchmark  use both RGB images and LiDAR point clouds whereas VoxelNet  uses only LiDAR.



We present several 3D detection examples in Figure~\ref{fig:Qualitative_demo}. For better visualization  3D boxes detected using LiDAR are projected on to the RGB images. As shown, VoxelNet provides highly accurate 3D bounding boxes in all categories.


The inference time for the VoxelNet is 225ms where the voxel input feature computation takes 5ms, feature learning net takes 20ms, convolutional middle layers take 170ms, and region proposal net takes 30ms on a TitanX GPU and 1.7Ghz CPU.

\begin{table}[!t]
\centering
\scalebox{0.9}{
\begin{tabular}{| c | c | c | c |}
\hline
{\bf Benchmark} & {\bf Easy} & {\bf Moderate} & {\bf Hard} \\ 
\hline
Car (3D Detection) & 77.47  & 65.11  & 57.73 \\
Car (Bird's Eye View) & 89.35  & 79.26  & 77.39 \\
\hline
Pedestrian (3D Detection) & 39.48  & 33.69  & 31.51 \\
Pedestrian (Bird's Eye View) & 46.13  & 40.74  & 38.11 \\
\hline
Cyclist (3D Detection) & 61.22  & 48.36  & 44.37 \\
Cyclist (Bird's Eye View) & 66.70  & 54.76  & 50.55  \\ 
\hline
\end{tabular}
}
\caption{Performance evaluation on KITTI test set.}
\label{table:test_set_result}
\end{table}

\section{Conclusion}
Most existing methods in LiDAR-based 3D detection rely on hand-crafted feature representations, for example, a bird's eye view projection. In this paper, we remove the bottleneck of manual feature engineering and propose VoxelNet, a novel end-to-end trainable deep architecture for point cloud based 3D detection. Our approach can operate directly on sparse 3D points and capture 3D shape information effectively. We also present an efficient implementation of VoxelNet that benefits from point cloud sparsity and parallel processing on a voxel grid. Our experiments on the KITTI car detection task show that VoxelNet outperforms state-of-the-art LiDAR based 3D detection methods by a large margin. On more challenging tasks, such as 3D detection of pedestrians and cyclists, VoxelNet also demonstrates encouraging results showing that it provides a better 3D representation. Future work includes extending VoxelNet for joint LiDAR and image based end-to-end 3D detection to further improve detection and localization accuracy.




\paragraph{Acknowledgement:} 

We are grateful to our colleagues Russ Webb, Barry Theobald, and Jerremy Holland for their valuable input.

{\small
\bibliographystyle{ieee}
\bibliography{refb}
}

\end{document}